\documentclass[preprint,12pt]{elsarticle}

\usepackage{subcaption}
\usepackage{lineno,hyperref}
\usepackage{natbib}
\usepackage{geometry}
\usepackage{fleqn}
\usepackage{graphicx}
\usepackage{newtxtext,newtxmath}
\usepackage{hyperref}
\usepackage{booktabs}
\usepackage{bm}
\usepackage{amsmath}
\usepackage{enumitem}
\usepackage{esvect}
\usepackage{siunitx}
\usepackage{soul,xcolor}

\modulolinenumbers[5]

\graphicspath{{./figures/}}
\journal{Elsevier}

\begin{document}

\begin{frontmatter}

\title{\large{Predicting Critical Heat Flux with Uncertainty Quantification and Domain Generalization Using Conditional Variational Autoencoders and Deep Neural Networks}}

\author[NCSU]{Farah Alsafadi}
\author[NCSU]{Aidan Furlong}
\author[NCSU]{Xu Wu\corref{mycorrespondingauthor}}
\cortext[mycorrespondingauthor]{Corresponding author}
\ead{xwu27@ncsu.edu}

\address[NCSU]{Department of Nuclear Engineering, North Carolina State University    \\ 
	Burlington Engineering Laboratories, 2500 Stinson Drive, Raleigh, NC 27695 \\}

\begin{abstract}
Deep generative models (DGMs) can generate synthetic data samples that closely resemble the original dataset, addressing data scarcity. In this work, we developed a conditional variational autoencoder (CVAE) to augment critical heat flux (CHF) data used for the 2006 Groeneveld lookup table. To compare with traditional methods, a fine-tuned deep neural network (DNN) regression model was evaluated on the same dataset. Both models achieved small mean absolute relative errors, with the CVAE showing more favorable results. Uncertainty quantification (UQ) was performed using repeated CVAE sampling and DNN ensembling. The DNN ensemble improved performance over the baseline, while the CVAE maintained consistent results with less variability and higher confidence. Both models achieved small errors inside and outside the training domain, with slightly larger errors outside. Overall, the CVAE performed better than the DNN in predicting CHF and exhibited better uncertainty behavior.
\end{abstract}


\end{frontmatter}

\newpage
\section{Introduction}

Deep generative learning \cite{ruthotto2021introduction} \cite{salakhutdinov2015learning} is a branch of deep learning that utilizes unsupervised learning to learn complex data distributions. It is capable of capturing the underlying structures and patterns within the training data. The learned data distributions can be leveraged to generate synthetic data samples that closely resemble the original training dataset for data augmentation. Since most data-driven machine learning (ML) models rely on ``big data'' to achieve a favorable accuracy, expanding smaller datasets enables better performance of the data-driven ML models trained with the expanded datasets. This is particularly beneficial in addressing the ``data scarcity'' challenge in fields like nuclear engineering, where limited data availability restricts the application of ML models in many problems.
Deep generative models (DGMs) offer a promising solution to the data scarcity issue. By utilizing well-trained DGM, one can significantly expand an existing dataset. Among the most widely used DGMs are generative adversarial networks (GANs) \cite{goodfellow2014generative}, variational autoencoders (VAEs) \cite{kingma2013auto}, normalizing flows \cite{rezende2015variational}, and diffusion models \cite{sohl2015deep}. These models adopt different approaches in learning the training data distribution and in generating new synthetic data.

VAEs, introduced by Kingma and Welling \cite{kingma2013auto}, adopt a unique approach to learn the underlying data distribution using variational inference \cite{blei2017variational}. A VAE model, just like a traditional autoencoder model, consists of three main components: the \textit{encoder}, the \textit{latent space}, and the \textit{decoder}. VAEs differ from traditional autoencoders in the encoding process. Instead of encoding data as deterministic values in the latent space, VAEs encode data as distributions, making them well-suited for data augmentation because one can simply sample from the latent space distributions and obtain new samples through the decoding process. 

VAEs have demonstrated successes in data augmentation across various domains, including acoustic modeling \cite{nishizaki2017data} and clinical studies \cite{papadopoulos2023variational}. They were also used to address the data scarcity challenge in healthcare domain by augmenting the Gram-stained smear images dataset, which improved the classification accuracy of the bacteria detection framework \cite{shwetha2024data}. In the realm of the industrial internet of things, VAEs were employed to address data imbalance, significantly improving the Macro-F1-scores of deep-learning-based intrusion detection systems \cite{liu2022intrusion}. Additionally, VAEs prove effective in enhancing product quality prediction through the generation of artificial quality values for training \cite{lee2023developing}. They are also frequently applied in computer vision tasks, such as generating static images \cite{walker2016uncertain} and enhancing image super resolution \cite{sonderby2016amortised}. Additionally, VAEs were used to generate synthetic vortex-induced vibrations data that was employed to train a transformer model for forecasting vibrations in time-space using sparse observations \cite{mentzelopoulos2024variational}. They were also utilized to generate temperature, velocity, and species mass fraction predictions within a computational fluid dynamics (CFD) data-driven surrogate model, allowing for predicting CFD data fields with reasonable accuracy \cite{laubscher2020application}. In nuclear engineering, a convolutional variational autoencoding gradient-penalty Wasserstein generative adversarial network with random forest (CVGR) was proposed to mitigate imbalance data problem in fault diagnosis of nuclear power plants \cite{guo2024imbalanced}. 

However, a basic VAE model can only generate synthetic samples \textit{randomly} instead of producing specific data instances, for example, data samples at conditions and domains desired by the user. To address this limitation, the conditional VAE (CVAE) model was proposed \cite{sohn2015learning}. CVAEs operate similarly to VAEs but utilize additional data for conditioning. During the training phase, instead of solely relying on the training data, CVAEs incorporate labels alongside the data to train both the encoder and the decoder. This incorporation of labels enables the decoder to learn how to generate data that corresponds to specific labels, allowing for targeted data generation. CVAEs have been applied in various domains, such as enhancing spectral data augmentation in practical spectroscopy measurements with limited labeled samples \cite{mu2022developing} and improving few-shot classification tasks \cite{zhang2021dizygotic}. They were also applied in energy systems fault detection and diagnosis, where CVAE was combined with GAN to address the issue of imbalanced samples by generating synthetic fault samples to balance the training dataset \cite{ruan2024fault}.
However, their application in nuclear engineering datasets has been rare. In a previous study, we investigated the effectiveness of CVAEs among other DGMs for data augmentation of void fraction simulations \cite{alsafadi2023deepNED}, demonstrating that CVAEs have a good potential in this field. 

In this work, we will present a detailed investigation of using CVAE models to augment an important set of experimental data in nuclear engineering thermal-hydraulics (TH), the critical heat flux (CHF). 
Critical boiling transition is a transition from a boiling flow regime that has a higher heat transfer rate to a flow regime that has a significantly lower heat transfer rate, which may result in fuel damage. Therefore, CHF has been one of the most concerned nuclear reactor operational characteristics. The design, evaluation, licensing and reliable operation of innovative reactor designs, such as advanced light water-cooled reactors, require a very good understanding of the critical boiling transition behavior. However, there is a critical lack of experimental data across a wide range of flow boiling conditions for CHF due to the time-consuming and costly nature of such experiments.
Recently, a database used to develop the widely known 2006 Groeneveld CHF lookup table \cite{groeneveld20072006} was published by the U. S. Nuclear Regulatory Commission (NRC) \cite{groeneveld2019critical}. This database (hereafter referred as the ``NRC CHF dataset''), consisting of nearly 25,000 data points, is the largest known CHF dataset publicly available worldwide with measurements in vertical uniformly-heated water-cooled cylindrical tubes. 

In this study, we develop a CVAE-based DGM for data augmentation of the NRC CHF dataset. One may wonder that $\sim$ 25,000 CHF data points seem to be a good amount of training data for ML tasks, why is DGM necessary in this case? We argue that after considering the wide ranges of reactor operating conditions, geometries and reactor types, the CHF data points are rather scarce and unbalanced, making data augmentation necessary, especially for heavily parameterized predictive ML models such as neural networks.
In order to assess the performance of the CVAE generative model, we directly compared it with a regular deep neural network (DNN) model. DNN models have proven to be powerful for regression tasks by directly learning the mapping between inputs and outputs in the training data. Both the CVAE and fine-tuned DNN models were trained and tested using the same datasets to generate CHF data under the same specified conditions, with the results subsequently compared. Uncertainty quantification (UQ) analysis was then applied via sampling of the CVAE model and ensembling of the DNN model. Domain generalization analysis was also performed for both models to assess their ability in extrapolating to new, unseen domains.

Several metrics were applied to assess the performance of the models. Comparison with true holdout values showed close alignment, indicating a low level of error in the generated and predicted CHF values. UQ results showed that the CVAE model has very small uncertainties in the generation of CHF values, whereas higher values were observed for the DNN model. Both the CVAE and DNN models showed similar behavior in their ability to generalize beyond their training domain, achieving small mean absolute relative errors. Both models have proven to be reliable in generating and predicting CHF values outside the training domain. The mean relative absolute errors were slightly smaller when predicting within the training domain, compared to predicting outside the training domain. The findings from this work show that the CVAE and DNN models were successful at generating and predicting CHF values with a high degree of accuracy, with the CVAE demonstrating a more favorable error behavior.  

The remainder of this paper is structured as follows: in Section \ref{sec:CHF_dataset}, we present the training data details. Section \ref{sec:methods} introduces the methodologies for the CVAE generative model and DNN regression model. The models' specifications along with UQ and domain generalization methods are presented in Section \ref{sec:ML_models}. The results of this work are presented in Section \ref{sec:results}, including uncertainty estimates and model domain generalization. Further discussions on this work are provided in Section \ref{sec:discussions}. Finally, our findings and conclusions are summarized in Section \ref{sec:conclusions}.

\section{The CHF Experimental Dataset}
\label{sec:CHF_dataset}

\begin{figure}[!ht]
    \centering
    \includegraphics[width=\textwidth]{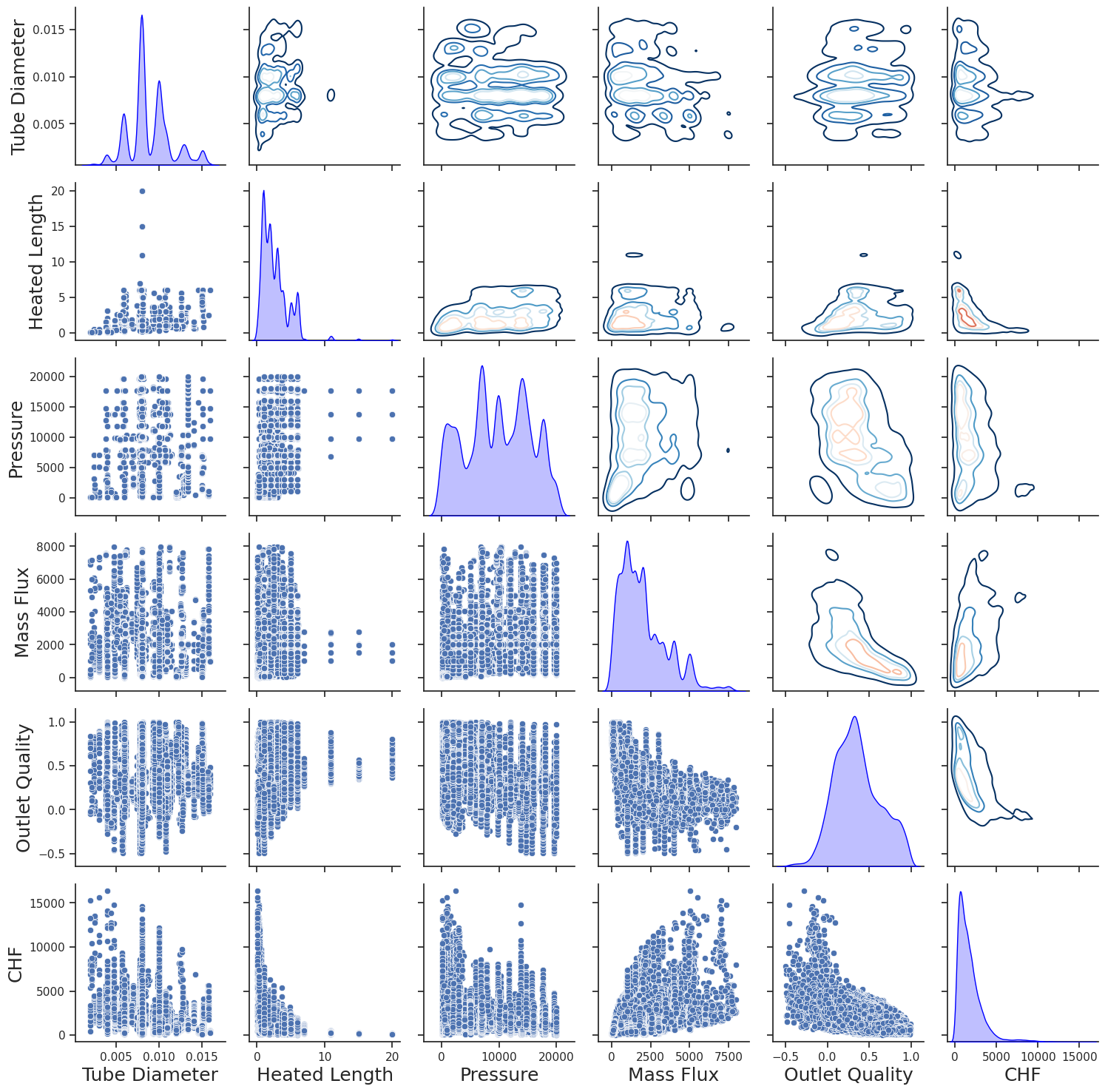}
    \caption{The distributions and correlations of the TH parameters and CHF values in the NRC CHF dataset.}
    \label{fig:CHF-data}
\end{figure}

CHF, also known as departure from nucleate boiling, is a critical phenomenon in heat transfer. It occurs when a heated surface reaches a point where it can no longer efficiently transfer heat to the surrounding fluid. In nuclear reactors, exceeding the CHF limit could potentially lead to fuel rod failure, which is a significant safety concern especially for pressurized water reactors. Therefore, it is essential to limit the heat flux of the fuel rods to a value below the CHF threshold. The collection of CHF measurement data can be challenging due to the nature of the experiments. Recently, the US NRC has published the largest known CHF dataset publicly available \cite{groeneveld2019critical}. It consists of nearly 25,000 CHF measurements in vertical uniformly-heated water-cooled cylindrical tubes, gathered over a span of 60 years from 59 different sources. In these experiments, CHF values were measured at various TH initial/boundary conditions: pressure ($P$), mass flux ($G$), inlet temperature ($T_{\text{in}}$), as well as geometrical parameters like test section diameter ($D$), and heated length ($L$). Additionally, the dataset contains calculated parameters derived from measurements and water properties, including outlet equilibrium quality ($x_\text{e}$), and inlet enthalpy ($\Delta h_{\text{in}}$). In this study, we selected $P$, $G$, $D$, $L$ and $x_\text{e}$ as the key TH parameters, along with  CHF for the analysis. The distributions of the values of these parameters, along with their pair-wise correlations, are illustrated in Figure \ref{fig:CHF-data}. 

Several ML approaches have been attempted in the prediction of CHF, including the use of DNNs and GANs. These studies have used a variety of datasets and input combinations including those not included in the public CHF database such as wall thickness \cite{nafey2009neural}, Reynolds' number \cite{xie2002hybrid}, and fluid densities \cite{lee2000correction}. Different ML methods have been proposed among these pre-existing works, ranging from purely DNN models to those incorporating physics-based knowledge \cite{zhao2020prediction}. The results of these approaches range widely, with reported root mean square error values ranging from $0.16\%$ \cite{vaziri2007critical} to $26.58\%$ \cite{lee2000correction} in different parameter configurations and methods. The authors are not aware of any presently-released studies attempting to use CVAEs. Another outstanding question is how the use of a far larger dataset such as the nearly $25{,}000$-entry public CHF compilation impacts the performance of DNN methods, and as such how it compares to the proposed CVAE generative model.

\section{Methodologies}
\label{sec:methods}

\subsection{DNNs} 

The use of DNNs has become widespread in multiple fields of nuclear engineering as a fast-and-accurate approach to create surrogate models, including in the prediction of CHF as discussed in Section \ref{sec:CHF_dataset}. DNNs are based on the transformation of a set of input values to produce a set of outputs (or single output) using a network of multiplying weights and additive biases. These are located on neurons, with an activation function applied to each of their outputs to introduce non-linearity. An objective function, typically the mean-squared error (\textit{MSE}), then evaluates how different the predicted outputs are from the true values. This value is then used to adjust the weights via backpropagation, which will minimize the objective function over a series of iterations through a section of the original dataset known as the training set. Once the surrogate model is fully trained, it is evaluated with a testing dataset that the model has never seen during training, providing a true estimate of the model’s ability to generalize to new data.

A DNN model is designed with weights organized into successive layers, with every neuron of one layer connected to every neuron in the subsequent layer. The depth of these models influence the complexity of information gained from the training set, with early layers extracting coarse features with the finer features extracted in deeper layers. Several hyperparameters influence the performance of a given model, such as the number of neurons per layer, the rate at which the weights are modified, and the choice of activation functions. As such, the optimization of these hyperparameters are necessary to ensure maximal performance of a given architecture. When a model is finished training and frozen, it is completely deterministic and will produce an identical output when given an identical input.

\subsection{VAEs and CVAEs}

\begin{figure}[!ht]
    \centering    \includegraphics[width = 0.7\textwidth]{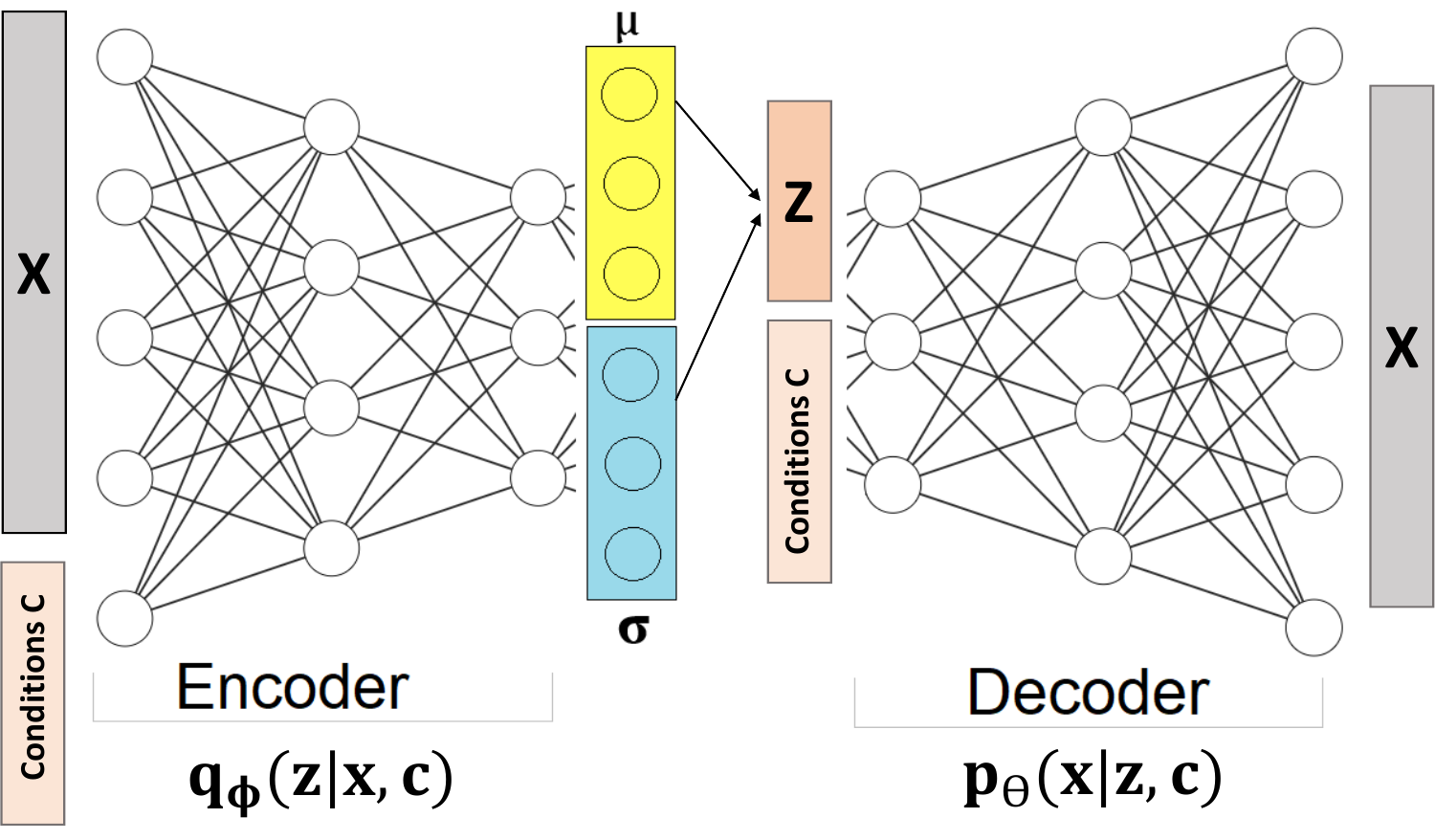}
    \caption{Illustration of the structure of a CVAE generative model.}
    \label{fig:CVAE}
\end{figure}

VAEs \cite{kingma2013auto} are a family of DGMs that was introduced to learn the data distribution uniquely through variational inference (VI) \cite{blei2017variational}. A VAE model consists of three main components: the encoder, latent space, and decoder, as shown in Figure \ref{fig:CVAE}. By passing the input to the encoder, it encodes it as a distribution in the latent space. Hence, the latent space consists of two vectors that represent the mean value and standard deviation of the encoded distribution. The decoder utilizes the encoded information to reconstruct the input by reversing the encoding process. This approach creates a structured latent space that can be effectively utilized for data generation. Once the model is trained, the decoder can be used to generate new samples by taking a vector from the latent space.  

This process does not allow for generating targeted samples, since the only input passed to the decoder is a random vector from the latent space. To enable targeted data generation, a CVAE is employed. CVAEs are a variant of VAEs that use labels or conditions for targeted data generation \cite{sohn2015learning}. Generating targeted data is achieved by passing specified conditions (vector $\mathbf{c}$) to the decoder as shown in Figure \ref{fig:CVAE}.

The loss function optimized during the training process is constructed by utilizing VI. Let $\mathbf{x}$ be a set of observed variables with a distribution denoted as $p(\mathbf{x})$, and $\mathbf{z}$ be a set of latent variables with a joint distribution $p(\mathbf{z},\mathbf{x})$. The conditional probability $p(\mathbf{z}|\mathbf{x})$ can be found by applying the Bayes' rule, $p(\mathbf{z}|\mathbf{x}) = p(\mathbf{x}|\mathbf{z}) p(\mathbf{z}) / p(\mathbf{x})$.  
Calculating $p(\mathbf{z}|\mathbf{x})$ is challenging due to the intractable integral $p(\mathbf{x}) = \int p(\mathbf{x}|\mathbf{z}) p(\mathbf{z}) d\mathbf{z}$. VI is used to approximate $p(\mathbf{z}|\mathbf{x})$ with a simpler distribution $q(\mathbf{z}|\mathbf{x})$, often chosen as Gaussian in practice. Afterwards, the differences between the two distributions is minimized via minimizing the Kullback–Leibler (KL) divergence. With this the loss function for VAEs can be written as follows: 
\begin{equation} \label{eqn:loss-function-derivation-vae}
\begin{aligned}
    \mathcal{L} (\bm{\theta},\bm{\phi})&=
   -\log ( p_{\bm{\theta}}(\mathbf{x}) ) + \mathcal{D}_{\text{KL}} ( q_{\bm{\phi}}(\mathbf{z}|\mathbf{x}) || p_{\bm{\theta}}(\mathbf{z}|\mathbf{x}) ) \\
    &=  - \mathbb{E}_{\mathbf{z} \sim q_{\bm{\phi}}(\mathbf{z}|\mathbf{x}) } \left[ \log(  p_{\bm{\theta}}(\mathbf{x}|\mathbf{z}) )\right] + \mathcal{D}_{\text{KL}} ( q_{\bm{\phi}}(\mathbf{z}|\mathbf{x}) || p_{\bm{\theta}}(\mathbf{z}) )
\end{aligned}
\end{equation}
Here, the approximation of the encoder is denoted by $q_{\bm{\phi}}(\mathbf{z}|\mathbf{x})$, and the decoder as $p_{\bm{\theta}}(\mathbf{x}|\mathbf{z})$. The parameters $\bm{\phi}$ and $\bm{\theta}$ represent the variational and generative parameters that undergo optimization during the training of the VAEs model. 

The loss function in Equation (\ref{eqn:loss-function-derivation-vae}) does not incorporate conditions, hence it is used to train the VAEs model. The CVAE's loss function can be constructed by conditioning the VAEs loss function on the conditions $\mathbf{c}$ as follows:
\begin{equation} \label{eqn:loss-function-cvae}
    \mathcal{L} (\bm{\theta},\bm{\phi}) = - \mathbb{E}_{\mathbf{z} \sim q_{\bm{\phi}}(\mathbf{z}|\mathbf{x},\mathbf{c}) } \left[ \log(  p_{\bm{\theta}}(\mathbf{x}|\mathbf{z},\mathbf{c}) )\right] + \mathcal{D}_{\text{KL}} ( q_{\bm{\phi}}(\mathbf{z}|\mathbf{x},\mathbf{c}) || p_{\bm{\theta}}(\mathbf{z}|\mathbf{c}) )
\end{equation}

Here, the KL divergence term is called the regularization term, which ensures that the learned distributions closely match the assumed Gaussian prior. The expectation term represents the reconstruction loss, which encourages the decoder to learn to reconstruct the data. Minimizing the loss function is equivalent to maximizing the lower bound of the probability of generating real data samples.

Generating new samples using CVAEs is done by passing a random vector from the latent space along with specified conditions to the decoder. This process yields different outputs, even when provided with the same conditions vector. This is due to the variability in the latent space inputs. The resulting variability can be used to estimate uncertainties in the model's predictions at any specific conditions \cite{gundersen2021semi}.

\section{The Generative and Predictive ML Models}
\label{sec:ML_models}

This work will make use of two ML models: a CVAE generative model and a DNN predictive model. In brief, the CVAE generative model tries to learn the underlying distribution of the NRC CHF dataset in order to generate synthetic samples at specific TH conditions. The DNN predictive model tries to learn the mapping between the inputs (TH conditions) and the output (CHF value) from the NRC CHF dataset in the format of regression. Once trained, the DNN model can also be evaluated at specific TH conditions, producing the corresponding CHF values. Their major differences can be summarized as: the CVAE model learns the underlying probabilistic distribution of the training dataset in an unsupervised way, while the DNN model constructs a black-box surrogate model of the training dataset without considering the data distribution in a supervised way.
If both models are good enough, the generated and predicted CHF values at the targeted TH conditions should be close to the real experimental data. In this section, we will present the details of the model architectures, as well as explain how to use UQ and domain generalization to compare the performance of these two ML models.

\subsection{The CVAE Generative Model}
\label{sec:ML_model_cvae}

The CVAE generative model will be used to approximate the underlying distribution of the NRC CHF dataset to generate new CHF values under specific TH conditions. The vector of TH conditions consists of five parameters: $P$, $G$, $D$, $L$, and $x_\text{e}$, as discussed in Section \ref{sec:CHF_dataset}. During the training process, these five parameters were provided to the model as conditions, that is, the vector $\mathbf{c}$ in Equation (\ref{eqn:loss-function-cvae}) and Figure \ref{fig:CVAE}, allowing it to learn how to generate CHF values under user desired conditions.
If a conventional VAE generative model were to be trained, one would not condition the model on the vector of TH parameters. In this case, the trained VAE model can only generate synthetic samples randomly, making it difficult to evaluate its performance by comparing with the real experimental data, since it is almost impossible for the random samples to have TH conditions that exactly match the real experimental data.

The model architecture was optimized by performing hyperparameter tuning for 400 iterations. Both the encoder and decoder networks were constructed with six fully connected layers. Training was conducted over around 360 epochs with a batch size of 52. Prior to training, the data underwent shuffling and standardization. The model was trained on 80\% of the NRC CHF dataset, while the remaining 20\% was reserved for the testing and validation, where the validation set was used for hyperparameter tuning. TensorFlow \cite{abadi2016tensorflow} was used to build the CVAE model. Once the model was trained, the decoder was used to generate CHF values under specific TH conditions in the testing dataset. Subsequently, the generated CHF values were compared with the corresponding experimental CHF values in the testing dataset to evaluate their accuracy.

\subsection{The DNN Predictive Model}

The DNN predictive model was also constructed in TensorFlow using eight hidden layers to transform the five input parameters listed in Section \ref{sec:ML_model_cvae}. The network depth, number of neurons per layer, activation functions, and learning rate were optimized using RayTune \cite{liaw2018tune} using 1,000 hyperparameter configurations in a random search. This tuning process was performed using the validation partition of the dataset. The training, validation, and testing datasets used between the CVAE and DNN were identical to ensure an accurate and fair comparison between the two methods. Once the hyperparameters were set, the model was trained for a total of 500 epochs with an exponential learning decay rate of 0.96. The fully trained network was then evaluated against the testing set using the set of metrics provided in Section \ref{sec:results}. The DNN model is much more straightforward to train since it only builds a black-box surrogate model of the training dataset without considering the data distribution. Our goal is to use such a fine tuned DNN model to ``benchmark'' the performance of the CVAE generative model using a comprehensive collection of metrics presented in the following sections.

\subsection{UQ of the ML Models}
\label{sec:ML_model_uq}

DNN models are often referred to as being ``black box'' due to their lack of interpretability as the dimensionality of the problem increases. In critical applications, such as safety analysis in a nuclear setting, it is imperative to quantify the level of uncertainty in a model's predictions to verify its trustworthiness. One such method commonly applied in DNN approaches is known as ensembling \cite{gawlikowski2023survey}, which relies on the idea that the consensus of several slightly-different networks will yield more reliable predictions than a single model. The multiple unique outputs per input set of an ensemble also allows for their treatment as a distribution, which can be assessed using traditional statistics such as standard deviation \cite{lakshminarayanan2017simple}. One of the simplest methods, albeit computationally intensive, to accomplish this is in the perturbation of the model's initializations.

The initialization of the weights and biases of a model, typically chosen at random, can significantly change its behavior. A given initialization will place the model at a specific starting point in the parameter space. The optimization process will then minimize the loss/objective function, potentially converging to a local minimum rather than a global minimum. This often leads to different weight configurations, which then leads to different outputs when comparing identical models with different initializations. This concept forms the basis of the initialization-based ensemble approach, where the random seeds are modified to produce slightly different models which produce slightly different outputs during testing \cite{lecun2015deep}. To implement this strategy, a total of 20 otherwise identical DNNs were trained, with different initialization random seeds, and their test outputs were collected. For each of the input entries in the testing set, there were now 20 unique CHF predictions, the mean and standard deviation of which were taken to obtain an estimation of the CHF prediction and its associated uncertainty.

Quantifying uncertainties in the generated samples by the CVAE model was done by utilizing the inherent variability in the generation process. Recall that the latent space in the CVAE model consists of two vectors that represent the mean value and standard deviation of the encoded distribution, which is usually assumed to be Gaussian. 
We first obtained 500 samples from the latent space distribution. Each of the latent vector samples is then combined with an input TH condition vector (the conditioning vector $\mathbf{c}$). The combined vector will be fed into the decoder to produce a CHF value. A different CHF value will be generated each time the decoder takes the same TH condition but different latent vector. This process was repeated for all TH conditions in the testing dataset. Afterward, the mean $\mu_{\text{samples}}$ and standard deviation $\sigma_{\text{samples}}$ were computed among the 500 samples for each TH condition. $\sigma_{\text{samples}}$ will be used to represent the uncertainty in the generated samples. It also indicates how sensitive the model predictions are to changes in the latent vector.

\subsection{Domain Generalization}

An important factor in ML model evaluation is assessing the model's ability to generalize to data outside the training domain. This involved assessing how accurately can the model extrapolate to new regions with unseen data. To this end, we employed the concept of a convex hull to evaluate both the CVAE and DNN models. The convex hull is the smallest convex set that includes all points in a given dataset.
In order to determine whether the point we are predicting at is inside or outside the training domain, we first construct a convex hull based on the training dataset. Afterward, the points in the testing dataset are classified into two categories; inside the convex hull or outside the convex hull. This is done by examining each testing data point to ascertain whether it falls within or outside the convex hull. If inside, the points are deemed to be within the domain covered by the training data; if outside, then the point is not represented in the space covered by the training data.

In this study, our testing dataset comprises 2,458 data points (10\% of the NRC CHF dataset). Of these, 2,307 points fall within the convex hull defined by the training dataset, while 151 points lie outside. Evaluating the two models' performance on generated/predicted CHF values for the 151 points outside the convex hull allows us to assess its capability for domain generalization.

\section{Results}
\label{sec:results}

\subsection{Results of CHF Generation and Prediction using CVAE and DNN}
\label{sec:results_nouq}

In this subsection, the generated CHF values using the CVAE model and the predicted CHF values using the DNN model at the TH conditions of the testing dataset will be directly compared to the true experimental CHF values. This is done by calculating the relative errors in the generated/predicted CHF values with respect to the true CHF values. Note that at this step we will not consider whether a sample in the testing dataset is inside or outside of the training domain.

\begin{figure}[!ht]
    \centering
    \begin{subfigure}{0.495\textwidth}
        \includegraphics[width=0.94\textwidth]{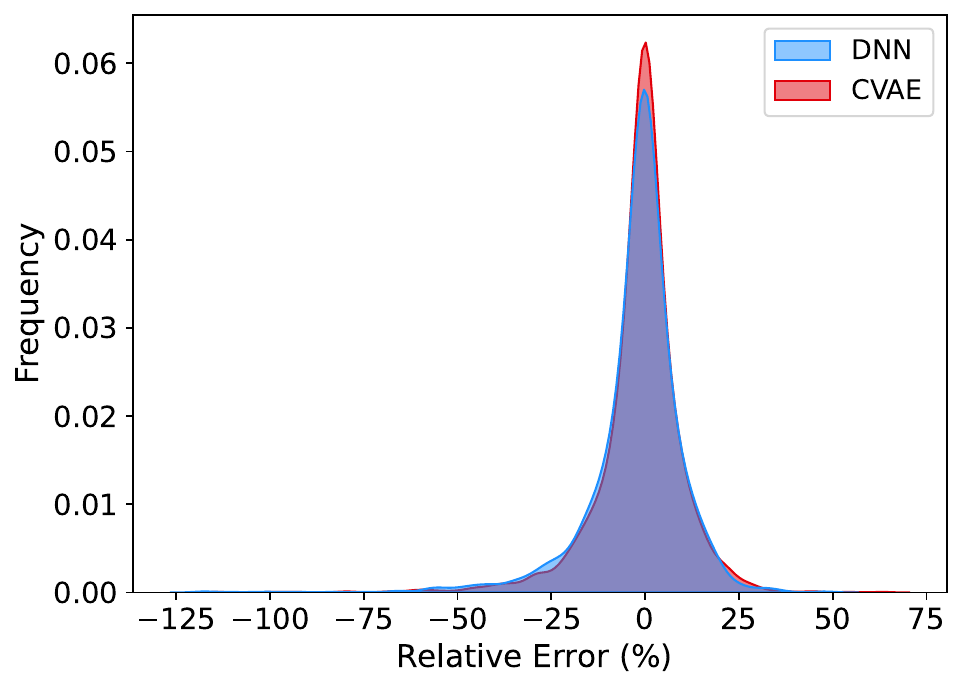}
        \caption{Relative error distribution}
        \label{fig:DNN-CVAEs-relative-error-distribution}
    \end{subfigure}
    \begin{subfigure}{0.495\textwidth}
        \includegraphics[width=0.99\textwidth]{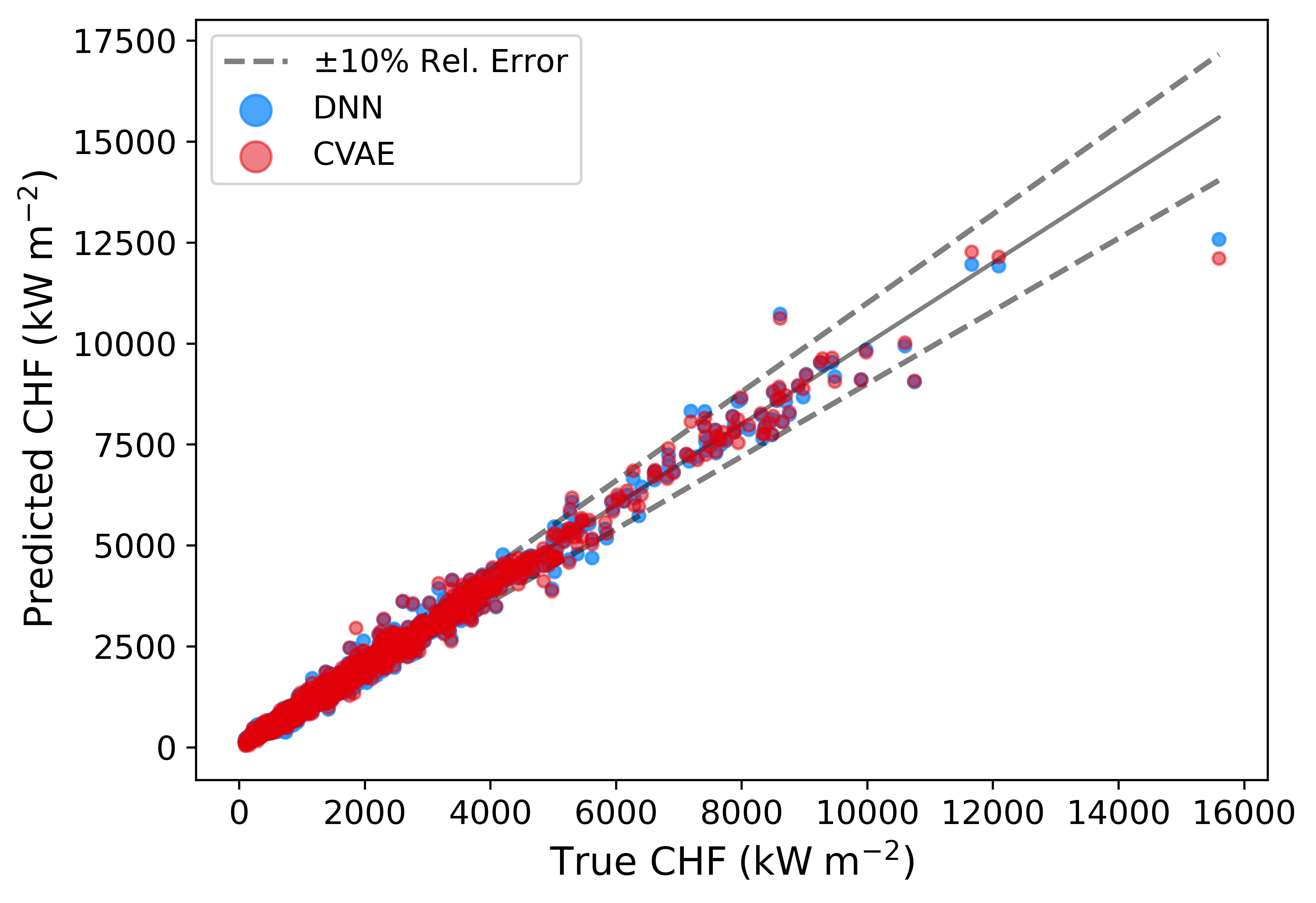}
        \caption{True vs. generated and predicted CHF values}
        \label{fig:DNN-CVAE-CHF-comparison-true-vs-predicted}
    \end{subfigure}
    \caption{Performance comparison of the DNN and CVAE models.}
    \label{fig:CVAEs-DNNs-predictions-comparison}
\end{figure}

Figure \ref{fig:DNN-CVAEs-relative-error-distribution} shows a comparison of the distributions of the relative errors from the CVAE and DNN models. The CVAE's error distribution is observed to have a slightly smaller mean value and standard deviation. Figure \ref{fig:DNN-CVAE-CHF-comparison-true-vs-predicted} shows a direct comparison of the true CHF values with those generated by the CVAE model and those predicted by the DNN model, with a $\pm 10\%$ error bounds. The results show a strong agreement with the true CHF values, with a slight increase in deviation observed after CHF value above 6,000 \si{\kilo\watt\per\square\meter}. It is also noticeable that there are few points having relative errors greater than 10\%.

Table \ref{table:metrics-RE-CVAEs-DNNs} includes statistical metrics calculated based on the absolute relative errors and the $R^2$ metric for both models. It provides the mean, maximum, and standard deviations of the absolute relative error values, along with the fraction of testing points resulting in an absolute relative error value greater than 10\%. The statistical metrics show that both models were able to predict CHF values accurately, with mean absolute relative error values of 7.292\% for the CVAE model and 8.455\% for the DNN model. The maximum error values were significantly higher than the mean values, however, these are observed at comparatively small CHF values, which can result in larger \textit{relative} errors despite small absolute differences. Notably, only 6.468\% of the testing data points had an absolute relative error greater than 25\% for the DNN model, compared to 4.353\% for the CVAE model.

    \begin{table}[ht!] 
	\normalsize
\captionsetup{justification=centering}
	\caption{Statistical metrics for the absolute relative errors.}
	\label{table:metrics-RE-CVAEs-DNNs}
	\centering
	\begin{tabular}{l c c}
        \toprule
        Metric & DNN & CVAE\\
        \midrule
        $\mu_{\text{error}}$ &8.455 \%  &   7.292 \% \\
        \hline
        $\text{Max}_{\text{error}}$&  94.40 \%  &  113.92 \% \\
        \hline
        $\text{Std}_{\text{error}}$ & 10.607 \%  &  9.329 \%\\
        \hline
        $F_{\text{error}}>10\%$ & 26.89  \%  &  23.79 \% \\
        \hline
        $\text{R}^2$ &  0.9845& 0.9880 \\
        \bottomrule
	\end{tabular}
\end{table}

The standard deviation values for both models indicate that the errors are not widely spread from the mean, suggesting that most predictions are close to the actual CHF values. Furthermore, both networks are observed with an $R^2$ value above 0.98 indicating a very high degree of correlation between true and generated/predicted values. Overall, these results show that both models perform well in generating and predicting CHF values, with the CVAE having more favorable values across all error metrics, except for $\text{Max}_{error}$ which was observed to be slightly higher.

\begin{figure}[!ht]
    \centering
    \includegraphics[width=0.9\textwidth]{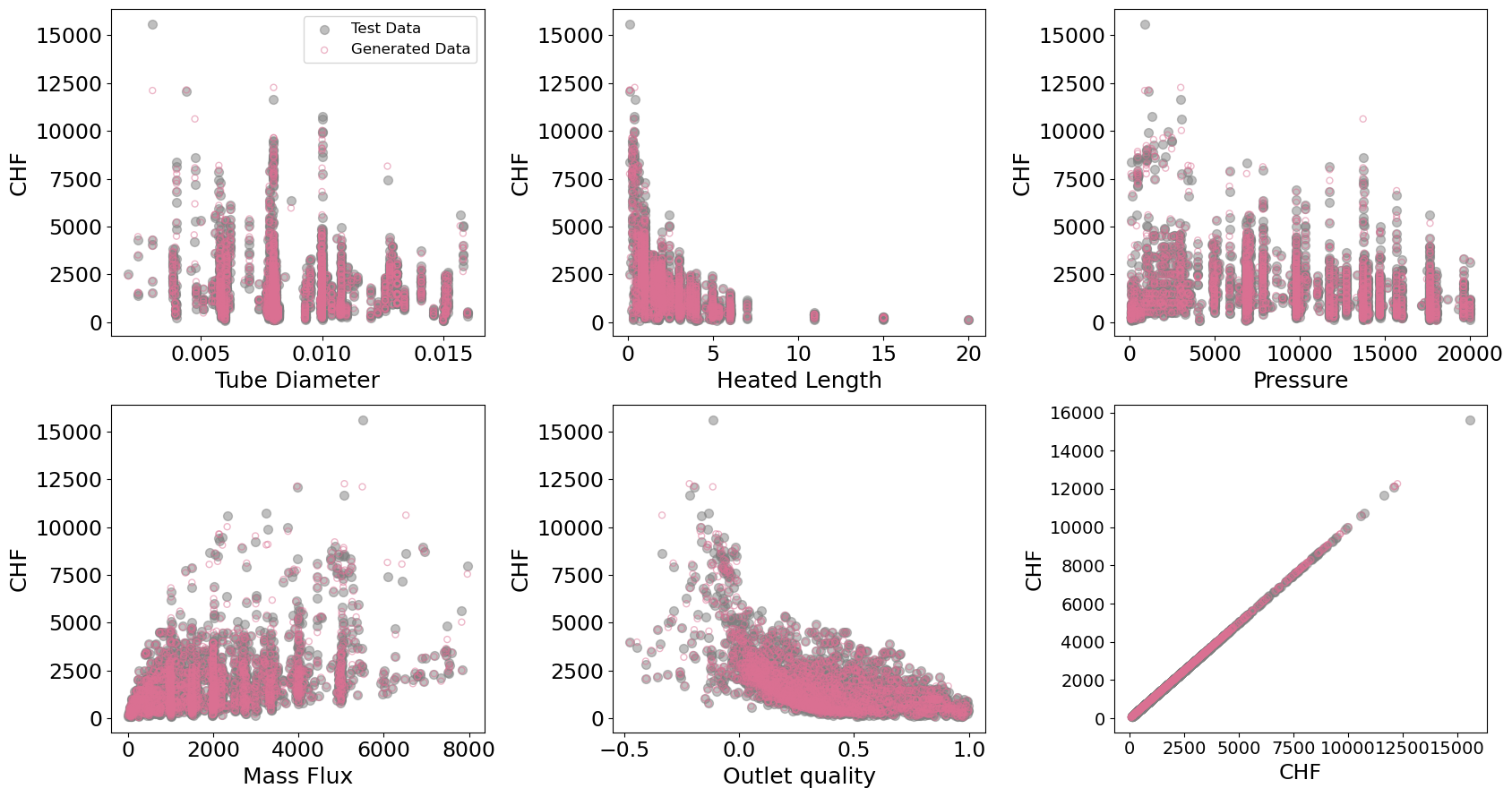 }
    \caption{Comparison of the CHF-TH-parameter pairwise correlations between the real data and the CVAE generated data.}
    \label{fig:CVAE-True-vs-Generated-Correlation-DLPGX}
\end{figure}

\begin{figure}[!ht]
    \centering
    \includegraphics[width=0.9\textwidth]{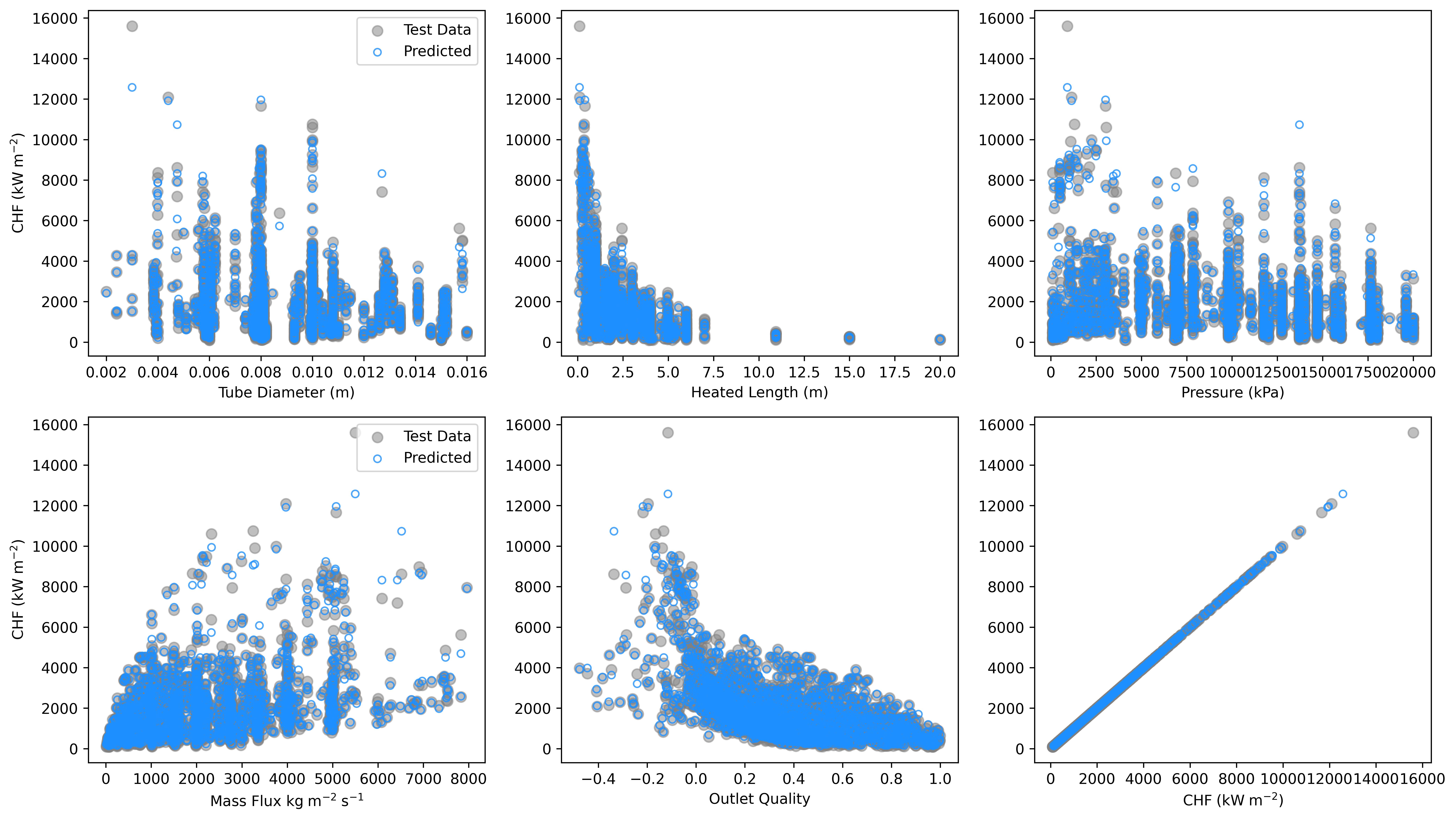}
    \caption{Comparison of the CHF-TH-parameter pairwise correlations between the real data and the DNN predicted data.}
    \label{fig:DNN-True-vs-predicted-Correlation-DLPGX}
\end{figure}

The last row in Figure \ref{fig:CHF-data} presents the pair-wise correlations between the measured CHF values and each of the five parameters in the vector of TH conditions. It is desirable to maintain such correlations in the generated and predicted CHF data. 
The correlations between the predicted and generated CHF values and the TH parameters in the testing dataset were compared with the real data to determine whether the models can learn and preserve these correlations. Figure \ref{fig:CVAE-True-vs-Generated-Correlation-DLPGX} compares the correlations of the TH parameters with the true CHF values and the generated CHF values using the CVAE model. Similarly, Figure \ref{fig:DNN-True-vs-predicted-Correlation-DLPGX} compares the correlations of the TH parameters with the true CHF values and the predicted CHF values using the DNN model. These figures also illustrate the TH parameter ranges in the testing dataset.

Figures \ref{fig:CVAE-True-vs-Generated-Correlation-DLPGX} and \ref{fig:DNN-True-vs-predicted-Correlation-DLPGX} show that the correlations between the TH parameters and the generated/predicted CHF values align closely with those found in the real data. Both models have consistent behavior in terms of the correlations. While largest relative absolute error is observed at smaller CHF values, the largest absolute deviations from the true values are observed at higher CHF values, which are associated with smaller values of $D$, $L$, and $P$, as well as subcooled coolant conditions, and high $G$ values. It is important to note that these regions contain fewer data points in both training and testing datasets, resulting in the model's deviations from the true values under these conditions. However, these deviations are not significant and do not heavily influence the error distributions, as these CHF conditions are less common in practical applications and thus in the CHF dataset.

\subsection{Results of UQ}

To quantify the model uncertainty, 500 unique samples were computed for each of the testing set's input entries using the trained CVAE model. From these, the means and standard deviations of these samples were taken for each of these input entries. The ensemble approach described in Section \ref{sec:ML_model_uq} was then implemented using $n = 20$ models each initialized with a different random seed. Each of these models were identical with the exception of the initialization. As with the CVAE model, the means and standard deviations were computed along the 500 samples for each input vector of the testing set. The standard deviations were then taken to compute the relative standard deviation using Equation \ref{eqn:rel_std}. This metric standardizes the uncertainty measure by expressing the standard deviation as a percentage of the mean, making it easier to interpret and to compare across different scales of the outputs.

\begin{equation} \label{eqn:rel_std}
    \text{Relative Std (\%)} = \frac{\sigma_{\text{samples}}}{\mu_{\text{samples}}} \times 100\%
\end{equation}

This process resulted in a relative standard deviation value for each of the output means, providing quantification of the model’s uncertainty. By comparing the relative standard deviations from the CVAE to those of the ensemble of DNNs, we can assess the effectiveness of each approach in capturing the uncertainty associated with the model predictions. This is done using the 6 metrics reported in Table \ref{table:metrics-RE-UQ-CVAEs-DNNs}, which also include the standard deviation of the error across all prediction means, as was done in Section \ref{sec:results_nouq}. In terms of error values, the CVAE achieves a smaller mean relative error of 7.26\% when compared to the 7.93\% of the CVAE. The standard deviation of this error distribution is also tighter in the CVAE's case, 9.28\% versus 10.42\%, but with generally comparable values when considering their small magnitudes. The DNN's maximum value of the error distribution, 118.50\%, is observed to be slightly larger than the CVAE's of 111.39\%. Both of these maximum values are located at relatively small CHF values near the origin.

\begin{table}[!ht] 
	\normalsize
	\captionsetup{justification=centering}
	\caption{Statistical metrics for the absolute relative errors with UQ.}
	\label{table:metrics-RE-UQ-CVAEs-DNNs}
	\centering
	\begin{tabular}{l c c}
        \toprule
        Metric & DNN & CVAE\\
        \midrule
        $\mu_{\text{error}}$ &  7.938 \%& 
           7.264 \%\\
        \hline
        $\text{Max}_{\text{error}}$& 118.50 \% & 111.39 \% \\
        \hline
       $\text{Std}_{\text{error}}$ & 10.424 \%&  9.2841\%\\
        \hline
        Mean relative std & 4.8240 \% &   0.6187 \%\\
        \hline
        Max relative std & 31.882 \%& 12.142 \%\\
        \hline
        $F_{\text{error}}>10\%$ & 26.078 \%  & 23.759 \% \\
        \bottomrule
	\end{tabular}
\end{table}

Now considering the distribution of relative standard deviations, the CVAE's results report significantly smaller mean and maximum relative Std values compared to the DNN's. The mean relative standard deviation of the DNN (4.82\%) is over seven times greater than that of the CVAE (0.6187\%), indicating that the CVAE is more confident in its predictions with less variability in its outputs. This is similarly mirrored in their maximum values, where the DNN has a value of nearly 31.88 \% with the CVAE reporting 12.14 \%. The final metric, the fraction of predictions with relative errors above 10\%, has comparable values between the two models. Considering the combination of the metrics in Table \ref{table:metrics-RE-UQ-CVAEs-DNNs}, the CVAE outputs are shown to have similar performance compared to a DNN ensemble while achieving more favorable values across all error metrics and significantly smaller model uncertainties.

\subsection{Results of Domain Generalization Analysis}

\begin{figure}[!ht]
    \centering
    \begin{subfigure}{0.485\textwidth}
        \includegraphics[width=.95\textwidth]{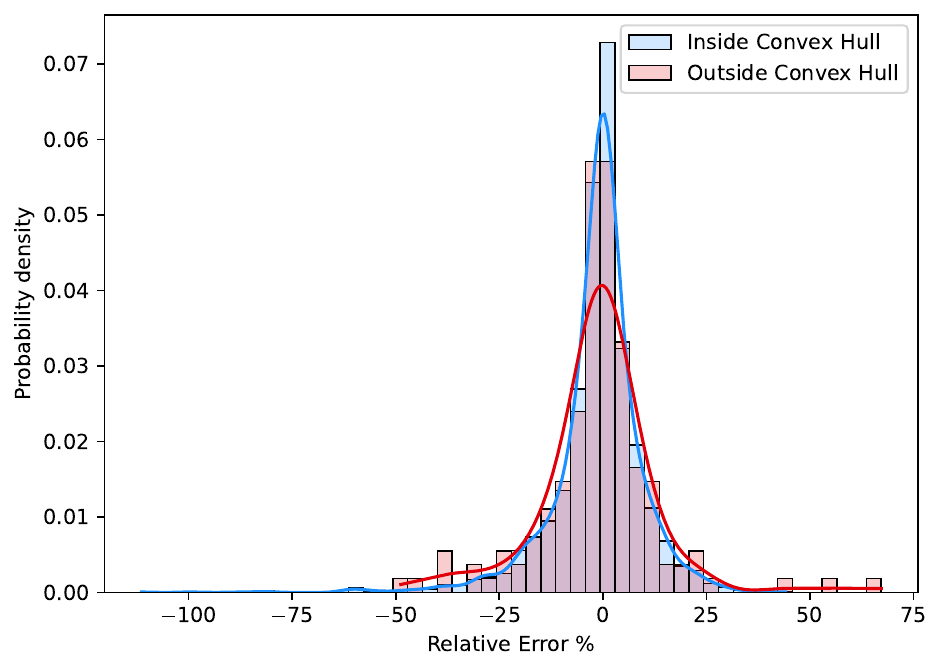}
        \caption{The CVAE generative model.}
        \label{fig:convex_hull_cvae_re}
    \end{subfigure}
    \begin{subfigure}{0.485\textwidth}
        \includegraphics[width=.94\textwidth]{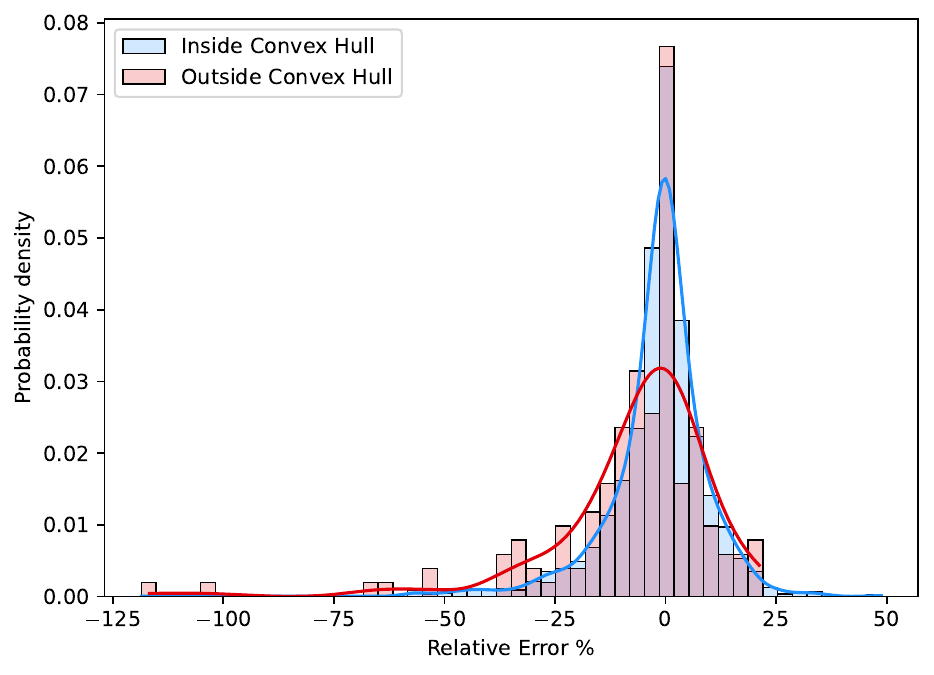}
        \caption{The DNN predictive model.}
        \label{fig:convex_hull_dnn_re}
    \end{subfigure}
    \caption{Distribution of the relative errors between $\mu_{\text{samples}}$ and the true CHF values, for testing points inside and outside the convex hull.}
    \label{fig:convex-hull-comparison}
\end{figure}

In this subsection, we will evaluate the performance of the CVAE generative model and DNN predictive model under conditions outside the training data domain. This was done by comparing the errors in their outputs based on conditions from both inside and outside the training data domain. The testing set used in this work consist of 2,458 data points, with 2,307 points falling within the training data domain and the remaining 151 points falling outside the domain, as determined by the convex hull defined by the training dataset. The domain generalization analysis was performed using the mean CHF values obtained from the ensemble of DNNs and, for the CVAE, the mean value from generating multiple outputs. The relative error between $\mu_{\text{samples}}$ and the true value was calculated for the points falling inside and outside the training domain. The distributions of the relative errors for these two subsets and two models are shown in Figure \ref{fig:convex-hull-comparison}.

As shown in Figure \ref{fig:convex-hull-comparison}, the error distributions for the two subsets (inside and outside the convex hull) behave similarly, with small deviations observed in the mean and standard deviation values. The maximum error values occur when predictions were made under conditions inside the training data domain. However, this does not necessarily indicate poorer model performance in this region. As mentioned earlier, these high relative error values result from small differences at low CHF values. This can be observed by comparing absolute errors (AE) inside and outside the convex hull. For the CVAE, the mean AE inside the convex hull was 89.641 \si{\kilo\watt\per\square\meter}, compared to 147.11 \si{\kilo\watt\per\square\meter} outside, while the maximum AE was 1100.12 \si{\kilo\watt\per\square\meter} inside and 3372.74 \si{\kilo\watt\per\square\meter} outside. This applies to both the CVAE and DNN models. The mean, maximum and standard deviation values were calculated for the absolute relative errors for both models and subsets. These values, along with the fraction of points with error values greater than 10\% in each subset are listed in Table \ref{table:CHF-metrics-convex-hull}. 

    \begin{table}[!ht]
    \normalsize
    \captionsetup{justification=centering}
    \caption{Statistics of the absolute relative errors for testing samples inside and outside the convex hull for the DNN and CVAE models.}
    \label{table:CHF-metrics-convex-hull}
    \centering
    \begin{tabular}{l r r}
        \toprule
        \textbf{Statistic} & \textbf{Inside training domain} & \textbf{Outside training domain} \\
        \midrule
        \multicolumn{3}{c}{\textbf{CVAE}} \\
        \midrule
        $\mu_{\text{error}}$ & 7.1442  \% &    9.0927 \%\\
        $\text{Max}_{\text{error}}$ &111.39 \% & 67.192 \% \\
        $\text{Std}_{\text{error}}$& 9.114 \% & 11.399 \%\\
        $F_{\text{error}}>10\%$ & 23.320 \%  & 30.263 \% \\
        \midrule
        \multicolumn{3}{c}{\textbf{DNN}} \\
        \midrule
        $\mu_{\text{error}}$ & 7.6951 \% &  11.625 \% \\
        $\text{Max}_{\text{error}}$ & 118.501 \% & 116.695 \% \\
        $\text{Std}_{\text{error}}$& 9.8447 \% & 16.503 \% \\
        $F_{\text{error}}>10\%$ & 25.498 \% & 34.868 \% \\
        \bottomrule
    \end{tabular}
\end{table}

Comparing the mean error inside and outside the training domain, one can observe that there is an increase in the mean relative error for both models. However, the increase is not significant, with mean error values remaining less than 12\% for both models and both subsets. A similar trend is observed for the standard deviation of the relative errors. This behavior is expected since extrapolation is being performed to unseen data domains. Notably, both models demonstrate favorable performance in predicting inside and outside the training domain. When predicting inside the training domain, fewer predictions resulted in errors greater than 10\%, while a slightly larger fraction of points exhibited errors above 10\% when predicting outside the training domain. This suggests that both models generalize well to new domains, maintaining acceptable error levels with mean errors below 12\%.

\section{Discussions}
\label{sec:discussions}

The difference between the CVAE generative model and the DNN predictive model lies in their distinct approaches: DNN learns the input-output mapping of the training data in a supervised way, while CVAE learns the underlying distribution of the training data in an unsupervised way and generates synthetic data by sampling from the learned distribution. Since CVAE does not learn direct mapping between the input and output, one would expect it to perform less accurate when compared to a fine-tuned DNN model that is trained for regression tasks. However, our findings showed that the CVAE generative model can be powerful in generating synthetic CHF values, even when compared to DNN. The CVAE model maintained favorable values in all error metrics when compared to the DNN model. 

The UQ analysis of the CVAE model showed that its generative predictions have small variations at a specific input. This is demonstrated by the very small mean relative standard deviations of its predictions, which is 0.6187\%. Additionally, changes in all metric values were very small when comparing the UQ and non-UQ resuts, as shown in Tables \ref{table:metrics-RE-CVAEs-DNNs} and \ref{table:metrics-RE-UQ-CVAEs-DNNs}. These indicate that the CVAE model has low sensitivity to the random samples from the latent vector during the generation process. The CVAE model is thus robust, consistently maintaining similar error behavior across multiple predictions. This consistency indicates that, although the model produces slightly different outputs each time, the predictions will remain within the quantified error ranges, ensuring reliable performance even with variations in the latent vectors.

On the other hand, quantifying uncertainties for the DNN model requires training multiple DNNs and measuring the variations between their predictions, which is more laborious than the CVAE model. The ensembled results showed improvement in all error metrics, except for the maximum error value, as shown in Tables \ref{table:metrics-RE-CVAEs-DNNs} and \ref{table:metrics-RE-UQ-CVAEs-DNNs}. 
This improvement can be attributed to the DNN model's high sensitivity to weight initialization during training. Averaging over an ensemble of DNNs can thus produce more accurate predictions. 
An ML model is generally expected to perform well for interpolation tasks, that is, when the model is applied to cases within the training domain. The ability of an ML model to extrapolate to new domains is important, especially if we aim to expand an existing dataset or test new experimental conditions. 

The domain generalization analysis showed that the CVAE and DNN models are reliable in generating and predicting CHF values outside the training domain. Both models showed consistent results when predicting both inside and outside the training domain. The mean relative absolute errors were smaller when predicting within the training domain, with a slight increase when predicting outside the training domain. However, both models maintained their favorable performance in predicting CHF values within and outside the domain. This indicates that both models were successful in extrapolating to new, unseen domains.

\section{Conclusions}
\label{sec:conclusions}

This study investigated the use of a Conditional Variational Autoencoders (CVAE) model in generation of critical heat flux (CHF) values and compared its performance with a traditional deep neural network (DNN) predictive model. This work was performed using the publicly-available CHF dataset used to construct the 2006 Groeneveld lookup table, consisting of nearly $25{,}000$ data points. Both the CVAE and DNN models were trained and tested using identical data partitions to ensure a direct comparison between them. The CVAE generative model demonstrated favorable test results across all error metrics when compared to the DNN predictive model.

Uncertainty quantification (UQ) was then performed for both models to evaluate their robustness and consistency. For the CVAE model, we generated 500 samples from the latent vector and combined them with each input entry in the testing dataset before entering the decoder to generate a CHF value. In this way, the CVAE model produces 500 CHF values for each data point in the testing dataset, which can be used to get the uncertainty in the CVAE generated data and provide insight into the variability of the model's predictions.
For the DNN model, an ensemble was created by training 20 models of identical architecture with different initializations of the weights and biases. Similar to the CVAE, this produces a set of 20 ``samples'' for each data point in the testing dataset. The means and relative standard deviations were then taken along the samples of each testing data point. 
The UQ analysis revealed that the CVAE model had significantly smaller relative standard deviations when compared to the DNN ensemble. The CVAE generative model thus has a higher confidence and lower variability in its outputs than the DNN predictive model.

In the subsequent domain generalization analysis, the sampled CVAE and DNN ensemble's outputs were then segmented into those inside and outside of the training set's domain to evaluate how their performance changes when extrapolating. Both CVAE and DNN models exhibited strong extrapolation capabilities to the data outside the training domain. Inside the training domain, the mean error for the CVAE was 7.14\%, and for DNN ensemble, it was 7.69\%. Outside the training domain, the mean error increased to 9.09\% for the CVAE model and 11.62\% for the DNN ensemble. Both models performed generally well when predicting/generating CHF both inside and outside the training domain. 

Overall, the CVAE generative model has proved to be an effective and reliable approach for generating CHF values, with low variability in predictions and consistent performance even when extrapolating to new domains. The UQ capabilities of CVAE are built-in with relative ease in creating multiple samples per input, especially when compared to the multimodel requirements of a DNN ensemble. These findings highlight the potential of CVAE for data augmentation applications in nuclear engineering and other fields requiring accurate and reliable predictions with quantified uncertainty. In future work, we intend to investigate the potential enhancement of CVAE performance in new domains through the use of transfer learning. It may also be worthwhile to explore the application of other deep generative techniques, such as diffusion models, for data augmentation.

\section*{Acknowledgement}

This work was funded by the U.S. Department of Energy (DOE) Office of Nuclear Energy Distinguished Early Career Program (DECP) under award number DE-NE0009467. Any opinions, findings, and conclusions or recommendations expressed in this paper are those of the authors and do not necessarily reflect the views of the U.S. DOE.

\newpage
\bibliography{./bibliography.bib}

\end{document}